# Teacher agency in the age of generative AI: towards a framework of hybrid intelligence for learning design

Thomas B. Frøsig
Université Côte d'Azur, France

https://orcid.org/0009-0002-2461-2645

Margarida Romero
Université Côte d'Azur, France

https://orcid.org/0000-0003-3356-8121

**Abstract**

Generative AI (genAI) is being used in education for different purposes. From the teachers' perspective, genAI can support activities such as learning design. However, there is a need to study the impact of genAI on the teachers' agency. While GenAI can support certain processes of idea generation and co-creation, GenAI has the potential to negatively affect professional agency due to teachers' limited power to (i) act, (ii) affect matters, and (iii) make decisions or choices, as well as the possibility to (iv) take a stance. Agency is identified in the learning sciences studies as being one of the factors in teachers' ability to trust AI. This paper aims to introduce a dual perspective. First, educational technology, as opposed to other computer-mediated communication (CMC) tools, has two distinctly different user groups and different user needs, in the form of learners and teachers, to cater for. Second, the design of educational technology often prioritises learner agency and engagement, thereby limiting the opportunities for teachers to influence the technology and take action. This study aims to analyse the way GenAI is influencing teachers' agency. After identifying the current limits of GenAI, a solution based on the combination of human intelligence and artificial intelligence through a hybrid intelligence approach is proposed. This combination opens up the discussion of a collaboration between teacher and genAI being able to open up new practices in learning design in which they HI support the extension of the teachers' activity.

*Keywords*: Hybrid Intelligence, Artificial Intelligence, Learning Design, Teachers' Agency, Educational Technology.

**Introduction**

Across history, different technologies have raised concerns about their impact on education, such as the introduction of the printing industry or mass media availability. People have been questioning the disruptive impact of educational technologies as a process of acculturation and integration since the last century. Cuban (1986) analysed a dataset of more than 60 years of empirical studies of educational technology integration and identified one pattern that might be of pivotal use in the debate surrounding generative AI (genAI) and its effect on the educational landscape. At the core of this pattern, Cuban describes a four-part cycle, which starts with the appearance of a new technology and subsequent predictions of changes in both student learning and teacher practices. According to Cuban, the first stage sparks enthusiasm and excitement among reformists, technologists, and policymakers. That leads the pattern into the second step of the cycle, which is where some academic studies focus on the effectiveness of innovation compared to traditional teaching practices. Within this second part of the cycle, scattered concerns and complaints from teachers and classroom observers about the usefulness of the technology in the classroom will also appear. The pattern then moves into the third part of the cycle, where the results of these concerns are documented in surveys as an infrequent and disappointing use of the technological innovation, thus establishing a low classroom usage of the tool that has been academically established as being effective. That leads to the fourth and last step of the cycle described by Cuban (1986), where a series of sharp critiques arise, blaming the low usage on teachers' resistance to change, their inertia, or conservatism. It is important to stress that while Cuban's cycle has been observed with educational technologies that are not genAI, there is a need to study the acculturation and integration of genAI in education from a non-deterministic perspective. In this paper, we aim to explore the emerging applications of human-AI collaboration in education, particularly emphasising the potential of hybrid intelligence to enhance the capabilities of both teachers and learners in supporting educational activities. As an illustration, genAI can be used to support socio-constructivist learning by supporting the Zone of Proximal Development (ZPD). The ZPD is defined by Vygotsky (1978) as "the distance between the actual developmental level as determined by independent problem solving and the level of potential development as determined under adult guidance or in collaboration with more capable peers" (p. 86), in which the genAI tool is considered the "more capable peer" in the human-AI interaction. For example, genAI such as ChatGPT can assist learners in interpreting historical documents by tailoring explanations to match their age and current level of understanding.

GenAI has gained widespread attention, and its potential is driving public discourse (Angwin, 2023) and its potential for education is being scrutinised by policymakers (Cardona et al., 2023), It does seem as if we are currently in the midst of the first part of Cubans' cycle. However, the current debate is not only showing generalised enthusiasm, but rather some cognitive dissonance by researchers in the learning sciences and educators, as pointed out by Mishra et al. (2023), stressing a polarisation on either positive or negative aspects of genAI technologies in education. According to Mishra and collaborators, the debate is see-sawing between genAI being the best/worst and whether we should embrace/reject it because it will transform our world into something amazing/horrible. In itself, this polarised debate would benefit from an increase in nuance, but moreover, it also suggests that the concerns and complaints have shifted away from appearing at the end of the scientific credibility part of the cycle to instead showing themselves during the exhilaration, or the very first part, of the cycle. To accommodate for this shift, the scope is widened to augment the Technological Pedagogical Content Knowledge framework (TPACK) (Mishra & Koehler, 2006) to extend the processes of educational technology integration identified by Cuban in the last decades. The TPACK model identifies technological knowledge (TK), pedagogical knowledge (PK), and content knowledge (CK) to understand the way teachers integrate technology into their practices. Additionally, contextual knowledge (XK) has been identified by Mishra (2019) as an important type of knowledge to be considered not only in the learning design but also in the regulation of the learning activities in the classroom. Specifically focusing on the updated domain of contextual knowledge (XK) which encourages a view shifting away from seeing teachers only as designers of learning activities in their classroom towards teachers also being innovation managers able to maneuver the domains of their organisation, policies and influence policymakers in order to create sustainable change (Mishra, 2019). This extension of the teachers' role requires teachers' agency but also an acculturation to the technological tools to support the change.

Through the perspective of XK it is possible to question the teacher's concerns and complaints regarding the usage of genAI in education. Teachers' criticism of genAI surfaces already during the first stages of genAI integration (Selwin, 2024), which corresponds to Cuban's exhilaration phase.

When teachers' expand their roles as decision-makers of the educational technologies they use in their classroom, there is a higher exposure to policies and organisational questions related to genAI.

What this paper seeks to contribute to the discussion, is the twofold notion that (i) educational technology, as opposed to other technology, has two distinctly different user groups and different user needs, in the form of learners and teachers, to cater for; and (ii) educational technology is, more often than not, designed with learner agency and engagement in mind, leaving limited possibilities for teachers to shape the technology and act (Frøsig, 2023). In order to evaluate these limited possibilities, this study aims to analyse the way genAI is influencing teachers' agency.

**Teachers' professional agency**

Professional agency, and specifically teacher agency, is a multidisciplinary concept covering four different subdisciplines, including (i) the social science traditions, (ii) the post-structural tradition, (iii) the socio-cultural approaches, and (iv) the identity and live-course approach (Eteläpelto et al., 2013, p.17). In their review of these different traditions, Eteläpelto and colleagues sustain that "professional agency is practiced when professional subjects and/or communities exert influence, make choices, and take stances in ways that affect their work and/or their professional identities" (Eteläpelto et al., 2013, p.17). In a qualitative meta-study investigating teacher agency within Finish Vocational Training, Vähäsantan (2015) sums up teacher agency as teachers having (i) the power to act, (ii) the power to affect matters, (iii) the power to make decisions and choices, and (iv) the power to take a stance.

**AI and teachers' agency in the literature**

Teacher's agency is known to be a key when trusting AI in education (Nazaretsky et al., 2022). The ability to exercise professional agency during AI integration enables teachers to mitigate misconceptions and false expectations about AI. This, in turn, promotes a more pragmatic and trustful approach to incorporating AI into specific teaching and learning activities, ultimately adding value from a human-centred perspective (Romero et al., 2024). Studies on teachers' agency and genAI are scarce, mostly focusing on the discrepancy between the actual output of an automated response system and the expected output by the teacher operating the system (Dietvorst et al., 2014) or, as in the case of Nazaretsky et al. (2022), the improvement of teachers' AI literacy.

Despite the limited focus on teachers' agency, this scarcity does not mean that genAI has not been a subject of study since it entered the public and educational domain in 2022, with different foci ranging from technology acceptance to the creative uses of genAI in education. Mishra and colleagues (2023) highlight the academic emphasis on topics like genAI's possibility to undermine academic integrity, thereby negatively impacting student learning. Or its inclination to 'hallucinate' and 'invent facts', triggering concerns about its contribution to the spreading of misinformation. Likewise, genAI's contribution to the widening of the digital divide between students who can afford access, and those who can't is highlighted, along with concerns about its built in biases caused by the vast amount of training data. This, in turn, ignites the ethical discourse surrounding copyright and the ownership of the training data. And while these topics focus more on the concerns, Mishra also highlights studies that advocate genAI's ability to positively enrich educational practices with topics such as personalised and differentiated learning, real time feedback, and offering assistance as a non human teacher and as an aid to human teachers (Mishra et al., 2023).

While the distribution of these topics depicts a polarised discourse about whether or not genAI can benefit in an educational context, Darvishi et al. (2024) utilised the fact that AI-powered tools are already being deployed in schools to automate and scaffold learning activities for students. In their study about the impact of AI assistance on student agency, they investigated 1625 students across 10 courses, presenting a result showing that students tend to rely on rather than learn from AI assistance. Stressing that while AI-powered learning technologies offer many advantages, their implementation should be approached carefully, striking a balance between AI assistance and fostering students' agency.

Expanding this notion from student agency into the domain of teacher agency would mean teachers relying on genAI to assist and generate output for them, be it in the form of teaching material, automated grading, or full lesson plans, would decrease their own agency, as they, after entering the prompt, do not have (i) the power to act, (ii) the power to affect matters, (iii) the power to make decisions and choices, and (iv) the power to take a stance, as genAI, by definition, generates its output autonomously. And even though such a practice would be in line with "teachers being the conduit for taking disciplinary knowledge and transforming it (using the right technology) for the benefit of learners and their educational development" (Mishra et al., 2023, p.11), the very nature of genAI, its ability to offer expertise in a wide array of subjects, its lack of explainability leading to experience a 'black box' when generating output, the ability to communicate socially in our language, and thus having the ability to behave as a real psychological 'other' calls into question many of the roles traditionally filled by teachers.

**Hybrid Intelligence as a potential for teachers' agency in learning design with genAI**

To resolve the conundrum of integrating genAI in education, without the risk of decreasing teacher agency, a solution might be found in the achievement of complex goals by combining human and artificial intelligence through a Hybrid Intelligence (HI) approach (Akata et al., 2020; Dellermann et al. 2019). It suggests going well beyond the current use by creating systems that operate as mixed teams, where humans and machines cooperate in synergy. In order to answer the question related to the way genAI is influencing teachers' agency, there is an opportunity to consider HI to offer a potential answer in a system designed to give teachers (i) the power to act, (ii) the power to affect matters, (iii) the power to make decisions and choices, and (iv) the power to take a stance. Such a system would be able to support teachers' learning design, without decreasing their agency. Mishra and colleagues (2023) summarise this notion in the process of moving teachers from just users or operators to co-creators, shaping, and being shaped by these technologies. The human-AI co-creation and the need to maintain a learner-centred education in the use of AI in education are essential not only from a human development perspective but also as an ethical requirement to avoid any type of automation that could harm the teachers' or learners' agency and competencies (Romero et al., 2024).

**Discussion**

Even though the idea of humans and technologies shaping each other in a continuous, dynamic co-constitution might sound far-fetched, the notion of a medium in itself being not neutral but rather an extension of man, is actually one of the foundational thoughts in medium theory (McLuhan, 1966). A popular example is the light bulb, which does not carry any content yet is a medium that has a vast social effect, which is that a light bulb enables people to create spaces during the nighttime that would otherwise be engulfed by darkness. In a similar line of thinking, genAI, when designed with HI and collaboration in mind, could open up new learning design practices that currently aren't accessible to teachers or could be more complex without genAI technologies. Making such a HI variant of genAI has the potential to transform teaching and learning practices in order to facilitate certain processes and to relocate the teachers' time and attention to the teacher-learner relationship.